\documentclass{article}
\PassOptionsToPackage{numbers, compress}{natbib}
\usepackage[preprint]{neurips_2020}
\usepackage[utf8]{inputenc} % allow utf-8 input
\usepackage[T1]{fontenc}    % use 8-bit T1 fonts
\usepackage{hyperref}       % hyperlinks
\usepackage{url,wrapfig}            % simple URL typesetting
\usepackage{enumitem,booktabs}       % professional-quality tables
\usepackage{amsfonts}       % blackboard math symbols
\usepackage{nicefrac}       % compact symbols for 1/2, etc.
\usepackage{microtype}      % microtypography
\usepackage{color}
\usepackage{amsmath}
\usepackage{graphicx}
\usepackage{caption}
\usepackage{subcaption}
\usepackage{textcomp}

\title{Frequency-compensated PINNs for Fluid-dynamic Design Problems}

\author{%
Tongtao Zhang \\
Siemens Corporate Technology \\
Princeton, NJ 08540 \\
\texttt{tongtao.zhang@siemens.com}
\And
Biswadip Dey\thanks{Corresponding author} \\
Siemens Corporate Technology \\
Princeton, NJ 08540 \\
\texttt{biswadip.dey@siemens.com}
\AND
Pratik Kakkar \\
Siemens Corporate Technology \\
Princeton, NJ 08540 \\
\texttt{pratik.kakkar@siemens.com}
\And
Arindam Dasgupta \\
Siemens Corporate Technology \\
Princeton, NJ 08540 \\
\texttt{arindam.dasgupta@siemens.com}
\And
Amit Chakraborty \\
Siemens Corporate Technology \\
Princeton, NJ 08540 \\
\texttt{amit.chakraborty@siemens.com}
}

\begin{document}
\maketitle

\begin{abstract}
  Incompressible fluid flow around a cylinder is one of the classical problems in fluid-dynamics with strong relevance with many real-world engineering problems, for example, design of offshore structures or design of a pin-fin heat exchanger. Thus learning a high-accuracy surrogate for this problem can demonstrate the efficacy of a novel machine learning approach. In this work, we propose a physics-informed neural network (PINN) architecture for learning the relationship between simulation output and the underlying geometry and boundary conditions. In addition to using a physics-based regularization term, the proposed approach also exploits the underlying physics to learn a set of Fourier features, i.e. frequency and phase offset parameters, and then use them for predicting flow velocity and pressure over the spatio-temporal domain. We demonstrate this approach by predicting simulation results over out of range time interval and for novel design conditions. Our results show that incorporation of Fourier features improves the generalization performance over both temporal domain and design space.
\end{abstract}

\section{Introduction}
The current approach for designing complex devices and systems, such as aero-dynamic surfaces and turbine components, typically involves an iterative interaction between design/operating space exploration and evaluation. However, high-fidelity fluid-dynamic simulations, which are necessary to evaluate the performance of design candidates under a variety of operating conditions, demand significant time and computational power. This limits the scope of the overall design optimization process and as a consequence,  may lead to sub-optimal design choices. Applying machine learning algorithms to develop a fast and accurate surrogate for predicting simulation outcomes has the potential to significantly accelerate design evaluations thereby generating improved design choices.

In recent years, deep neural networks and representation learning \cite{goodfellow2016deep} have improved significantly in accuracy and is widely-used in many application domains, such as image recognition \cite{he2016deep}, sequential decision making \cite{silver2017mastering}, and language comprehension \cite{devlin2018bert}. These approaches, especially supervised learning, leverage datasets of inputs (e.g. pictures of people, historical data of a region’s average weather data, etc.) and outputs (e.g. identity of the individuals in a picture, and weather forecast, respectively) to learn their functional relationship using backpropagation. As inference in a neural network involves a single forward pass this provides an excellent opportunity to learn good-quality, low-cost surrogates for exploring the design spaces associated with fluid-dynamic problems.

To enable generalization beyond the training set, learning approaches incorporate appropriate inductive bias \citep{baxter2000model, haussler1988quantifying} and promote representations which are \emph{more plausible} in some sense. It typically manifests itself via a set of assumptions, which in turn can guide a learning algorithm to pick one hypothesis over another. The success in predicting an outcome for previously unseen data then depends on how well the inductive bias captures the ground reality. Inductive bias can be introduced as a prior in a Bayesian model, or via the choice of computational graphs and regularization terms in a neural network. In problems wherein laws of physics have a strong influence on the input-output relationship, generalization can be improved by leveraging underlying physics for defining the regularization term. 

In this work, we incorporate a physics-based regularization term so that a learned surrogate conforms to the underlying physics governed by Navier-Stokes equations. We develop a PINN framework to infer how the velocity and pressure fields associated with the flow of an incompressible fluid around a cylinder depend on the underlying geometry and boundary condition - in particular, the size of the cylinder and inlet velocity, respectively. In addition, to capture the periodic nature of the solution which is governed by the Strouhal number \citep{anderson1995computational}, the proposed approach learns a set of Fourier features and a phase offset parameter as functions of the underlying geometry and boundary conditions. After the PINN has been trained, simulation results, i.e. velocity and pressure fields, for new design choices can be inferred in a fast, computationally inexpensive way. Thus, by enabling high-throughput evaluation of potential design candidates, this proposed approach provides a means to achieve better, more efficient design solutions. The key contributions of this work are as follows:
\begin{itemize}[leftmargin=1.5em,noitemsep,topsep=-8pt]
    \item We introduce a Fourier Feature Mapping (FFM) subnetwork within a PINN framework to yield better predictions about fluid flow around a cylinder. The FFM subnetwork learns frequency and phase offset parameters as a function of cylinder shape and inlet velocity.
    \item Subsequently, we use the learned surrogate to predict simulation outputs for novel design conditions and demonstrate the improvement in its generalization performance.
\end{itemize}

\subsection*{Related Work}
\paragraph{ML-based Approaches for Fluid-dynamic Simulations:}
Use of machine learning algorithms in fluid-dynamic problems has drawn significant attention over the last few years. \cite{hennigh2017lat, wei2018learn} have shown that supervised learning, using large datasets of simulation results obtained from finite-element or finite-volume solvers, can build surrogates for predicting predict simulation results with high accuracy. \cite{jiang2020meshfreeflownet, nabian2020physics, raissi2019physics, wang2020towards} have demonstrated that ML-based approaches can predict simulation results in mesh-free manner and incorporation of physics-based regularization in these formulations improves the quality of results by a significant margin. In addition, supervised learning has also been used to guide the discretization process in a data-driven way \citep{bar2019learning} or to learn efficient iterative solvers \citep{hsieh2018learning}. On the other hand, alternative approaches \cite{dwivedi2019distributed, lu2019deepxde, nabian2019deep} based on self-supervised learning have also been proposed; they employ a neural network to approximate the solution of a differential equation and then use automatic differentiation to compute the loss function which is a quantitative measure on how well the dynamics (represented via a differential equation) and the initial/boundary conditions are enforced. As these approaches use the physics itself to generate training data, this line of of work completely avoids the computationally expensive process for generating simulation datasets. It has also been shown that such neural network based approximations for a class of quasilinear, parabolic partial differential equations can converge to their true solutions with arbitrary accuracy \citep{sirignano2018dgm}.
\paragraph{Frequency Bias in Neural Networks:}
Frequency bias in neural networks is a relatively well-studied problem, with a body of work focusing on the relationship between frequency components present in a function and the speed at which neural networks learn them. \citep{pmlr-v97-rahaman19a} has demonstrated that neural networks with ReLU activation favors functions with low frequency components. \citep{eldan2016power, NIPS2014_5422} have shown that deeper architectures are needed for a neural network to learn high-frequency functions. \citep{basri2020frequency, NIPS2019_8723} analyzed the learning dynamics in gradient descent and have shown that neural networks learn low frequency functions much faster than high frequency functions. On the other hand, since the seminal work by Rahimi and Recht \citep{rahimi2008random}, multiple work have focused on accelerating learning algorithms by mapping the input to an appropriate feature space. A recent work \citep{tancik2020fourfeat} has shown that by using Fourier features as inputs to a multi-layer perceptron (MLP) can improve its capability to learn high frequency functions. Another work \citep{sitzmann2020implicit} has shown that use of periodic activation functions (e.g., $\sin$) can provide an efficient tool to learn representations in a large class of problems.

\section{Frequency-compensated PINN for Fluid-dynamic Problems}
\subsection{Navier-Stokes Equations}
In this work, we consider a classical fluid-dynamics problem, namely a cylinder in the cross flow \citep{anderson1995computational}. In this flow configuration, an incompressible fluid passes around a cylinder. Then, by letting $u$, $v$ denote the horizontal and vertical components of the velocity field and $p$ denote the pressure field, the dynamics can be expressed as:
\begin{equation}
\label{eqn:Dynamics}
\left.
\begin{array}{rl}
\frac{\partial u}{\partial t} + u\frac{\partial u}{\partial x} + v\frac{\partial u}{\partial y} + \frac{\partial p}{\partial x} 
-\nu\left(\frac{\partial^2u}{\partial x^2} + \frac{\partial^2u}{\partial y^2}\right)
&= 
0
\\
\frac{\partial v}{\partial t} + u\frac{\partial v}{\partial x} + v\frac{\partial v}{\partial y} + \frac{\partial p}{\partial y} 
- \nu\left(\frac{\partial^2v}{\partial x^2} + \frac{\partial^2v}{\partial y^2}\right)
&= 
0
\\
\frac{\partial u}{\partial x} + \frac{\partial v}{\partial y}
&= 0
\end{array}
\quad \right\} \quad
\Psi(u,v,p) = 0,
\end{equation}
where $\nu$ denotes the kinematic viscosity of the fluid under consideration and the functional $\Psi$ describes the underlying partial differential equation. Moreover, by letting $d_y$ and $u_{inlet}$ denote the cylinder's diameter and inlet velocity, respectively, the Reynolds number for this flow can expressed as $\mathbf{Re} = (u_{inlet} * d_y)/\nu$. As we assume the kinematic viscosity to be $0.001$, the \textit{vortex shedding frequency} can be approximated as $0.21 * (1 - 21/\mathbf{Re})*(u_{inlet}/d_y)$ for the range of inlet velocity and cylinder diameter considered in this work \citep{sumer2006hydrodynamics}.
\subsection{PINN Architecture and the Fourier Component}
\begin{figure}[ht]
    \centering
    \includegraphics[width=\textwidth]{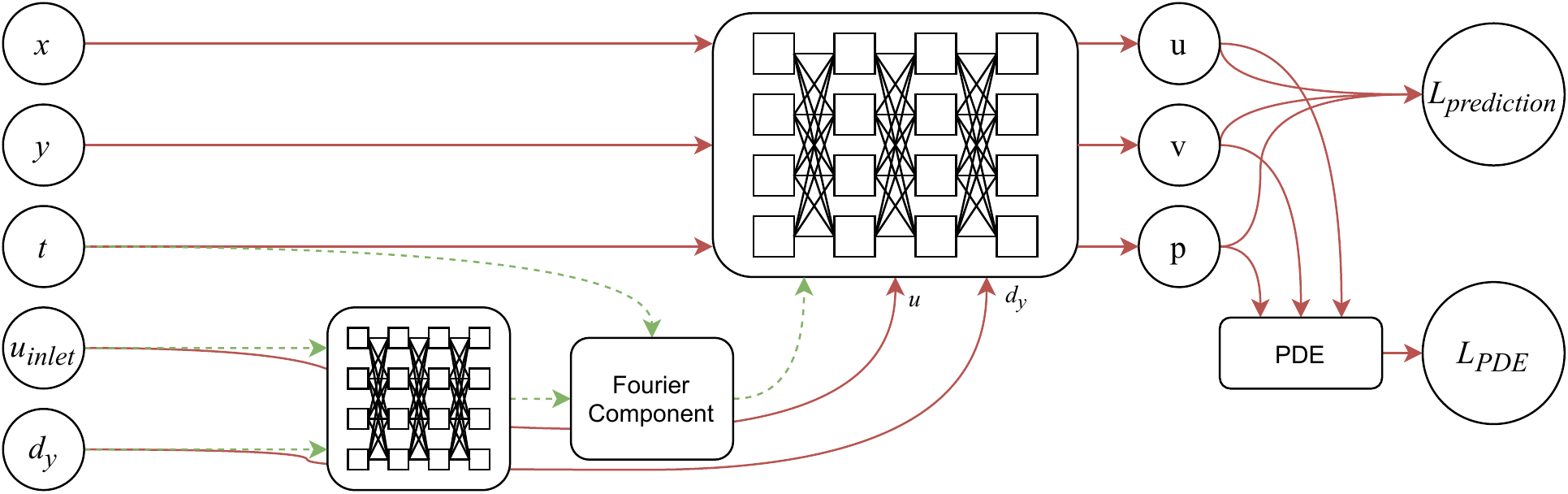}
    \caption{\small{A diagram demonstrating the proposed framework. Arrows with dotted lines denote the data flow with regard to Fourier Feature Mapping.}}
    \label{fig:network}
  \end{figure}

%\subsection{Neural Network Architecture}
%
Figure~\ref{fig:network} illustrates the PINN architecture that we propose to learn a surrogate for the aforementioned problem. The input to the network are as follows: the spatial coordinates inside the domain ($x,y$), the temporal dimension ($t$), the inlet velocity ($u_{inlet}$) and the diameter of the cylinder ($d_y$); and the PINN predicts the velocity ($u,v$) and pressure ($p$) of the flow.
  
As mentioned in the previous subsection, the velocity and pressure exhibit a periodic behavior along the temporal dimension. To capture this aspect, we use a set of Fourier features defined as
\begin{equation}
\label{eqn:fourier}
\textrm{Fourier Component} = [\sin(Ft+\phi), \cos(Ft+\phi)],
\end{equation}
and employ an MLP to learn the frequency ($F$) and phase shift ($\phi$) in the Fourier component as a function of $u_{inlet}$ and $d_y$. These features are then fed into a second MLP subnetwork, along with the spatio-temporal coordinates ($x,y,t$) and design specifications ($u_{inlet},d_y$). The output from this second subnetwork yield predictions on flow velocity and pressure. By letting $\hat{u}$, $\hat{v}$ and $\hat{p}$ denote the predicted velocity and pressure, respectively, we define the prediction error as:
\begin{equation}
    L_{prediction} = \frac{1}{n}\sum_{i=1}^{n}[(u_i-\hat{u}_i)^2 + (v_i-\hat{v}_i)^2 + (p_i-\hat{p}_i)^2],
\end{equation}
where $n$ is the training dataset size. In addition, to enforce that the predicted values conform to the underlying physics governed by \eqref{eqn:Dynamics}, we use the following regularization term
\begin{equation}
    L_{PDE} = \frac{1}{n}\sum_{i=1}^{n}\|\Psi(u_i,v_i,p_i)\|^2.
    \label{Loss-PDE}
\end{equation}
Finally we define the following loss function for the PINN
\begin{equation}
L = L_{predition} + \lambda L_{PDE},
\end{equation}
where the hyper-parameter $\lambda$ maintains a balance between prediction accuracy and regularization. 
\section{Experiment}
To evaluate performance of our proposed PINN framework, we train the network using simulation data corresponding to a handful of predefined geometry ($d_y$) and inlet velocity ($u_{inlet}$) combinations and use it to predict outputs for other combinations of geometry and inlet velocity values.

\subsection{Dataset}
\label{subsec:dataset}
\begin{wrapfigure}{r}{0.75\textwidth}
  \vspace{-1em}
  \begin{center}
    \includegraphics[width=0.72\textwidth]{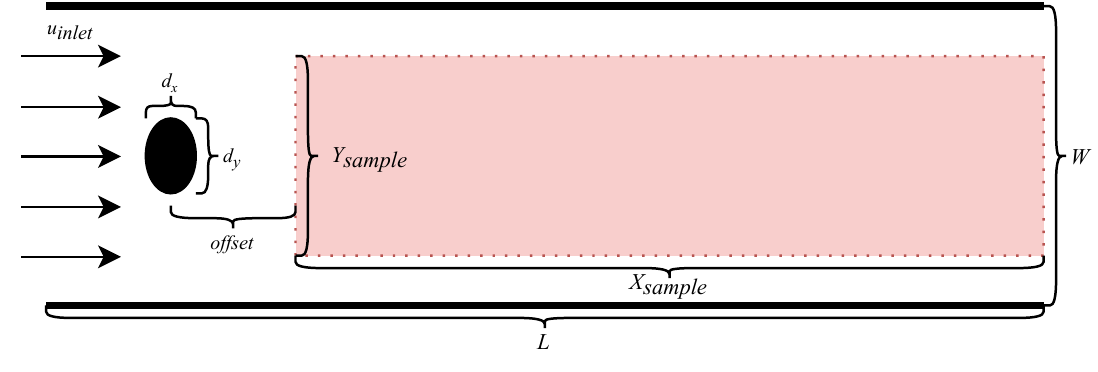}
  \end{center}
  \vspace{-1em}
  \caption{\small{Geometry of the flow configuration. The pink area highlights the region of interest which is aligned in the vertical middle and is located in the right of the cylinder after an \textit{offset} of $0.1$ from its center.}}
  \vspace{-1em}
  \label{fig:geometry}
\end{wrapfigure}

We use FeniCS \cite{AlnaesBlechta2015a,LoggMardalEtAl2012a}, a finite-element solver, to create a dataset for the flow around a cylinder problem, as illustrated in Figure~\ref{fig:geometry}. We have a rectangular channel with length $L=1.8$ and width $W=0.4$, and an elliptical cylinder with fixed horizontal diameter $d_x=0.1$ and vertical diameter 
$d_y$ is placed inside the channel. The cylinder is placed at the vertical middle point and $0.2$ from the inlet on the left. The region of interest is a rectangular region of length $X_{sample}=1.5$ and width $Y_{sample}=0.3$. We run FEniCS simulation for the time interval $[0s,6s]$ (with a time resolution $0.05s$), and then sample the velocity and pressure data for this region of interest. In this work, we train and evaluate the PINN predict the output within this region of interest (shown in pink in Figure~\ref{fig:geometry}).

In this work, we assume $u_{inlet}\in[0.8, 1.0]$ and $d_y\in[0.08,  0.11]$ and form the training set by running FEniCS simulation for the following 9 combinations ($0.8, 0.08$), ($0.8, 0.09$), ($0.8, 0.10$), ($0.9, 0.08$), ($0.9, 0.10$), ($0.9, 0.11$), ($1.0, 0.08$), ($1.0, 0.09$), and ($1.0, 0.11$) and then only taking the points from the time interval $[0s,5s]$ with the interval of $0.1s$. Also, we keep the result for the following combinations of $u_{inlet}$ and $d_y$ as the validation set - ($0.9, 0.09$) and ($0.9, 0.10$); although the combination ($0.9, 0.10$) appears in the both these sets they they do not overlap in the time direction. The rest of the points constitute the test set. With this, we have a dataset with training set of $517,350$ instances, validation set of $206,700$ instances and test set of $827,520$ instances\footnote{For further details such as data point distribution among the geometry settings, please read the supplementary materials}. If a geometry setting (i.e. a particular combination of $u_{inlet}$ and $d_y$) appears in training set, we name it ``Seen'', otherwise ``Unseen''.

\subsection{Parameters and Settings}
For the Fourier Feature Mapping (FFM) subnetwork, we use $3$ fully-connected (FC) layers with $120$ neurons on each layer and use \verb+Tanh+ activation function after the dropout layer. The final output size of the subnetwork is set as $10$, and the first $5$ outputs serve as frequency $F$ in the Fourier component \eqref{eqn:fourier} while the rest $5$ serve as phase shift $\phi$.

For the second MLP subnetwork, we set up a network consisting of $10$ FC layers with $120$ neurons on each layer. The activation function for each FC layer is also chosen as \verb+Tanh+.

In each step, besides the minibatches from the training set, we also randomly sample points from the recatangular region and we calculate the PDE loss \eqref{Loss-PDE} for those randomly sampled points so that the learned surrogate conforms to \eqref{eqn:Dynamics}. The partial deriverative equations are implemented using \verb+autograd+ toolbox from PyTorch~\cite{NEURIPS2019_9015}. The random points are drawn in the following domains: $x \in [0, 1.5], y \in [0, 0.3], t \in [0, 5], u_{inlet} \in [0.8, 1]$ and $d_y \in [0.08, 0.11]$.

We use Adam Optimizer~\cite{kingma2014adam} to train the whole neural network with a learning rate of $0.001$. The default run is $20,000$ epoches with the minibatch size of 32768, and we adopt early-stopping strategy if the validation loss does not reduce.

\subsection{Results}
\label{sec:result}
We show the errors and illustrate visualization output from our proposed framework given different geometry and component settings. Table~\ref{tab:loss_values} includes the MSE loss values at all the geometry and component settings which will be analyzed through the whole subsection.

\begin{table}[ht]
  \centering
  \begin{tabular}{ccccc}
      \toprule
      & Seen/Covered & Unseen/Covered & Seen/Uncovered & Unseen/Uncovered\\
      \midrule
      Full Component & \textbf{1.09$\times$10$^{-4}$} & \textbf{0.0269} & \textbf{0.0727} & \textbf{0.112}\\
      No-FFM& $1.73\times10^{-4}$ & 0.347 & 0.0973 & 0.163\\
      Strong-Reg& $8.47\times10^{-4}$ & 0.0589 & 0.0453 & 0.0986\\
      No-Reg/Overfit & $2.51\times10^{-3}$& 0.0431 & 0.0649 & 0.126\\
      \bottomrule
    \end{tabular}
  \caption{Mean square error values of the testing sets and different experiment settings which will be discussed through Section~\ref{sec:result}. ``Seen/Unseen'' denotes whether the geometry settings appear in the training set (the data points in training and test set never overlap), ``Covered/Uncovered'' denotes whether the data is drawn from $t\in[0,5]$ span.}
  \label{tab:loss_values}
\end{table}

\begin{figure}[ht]
  \centering
  \begin{subfigure}[b]{0.8\textwidth}
  \centering
  \includegraphics[width=\textwidth]{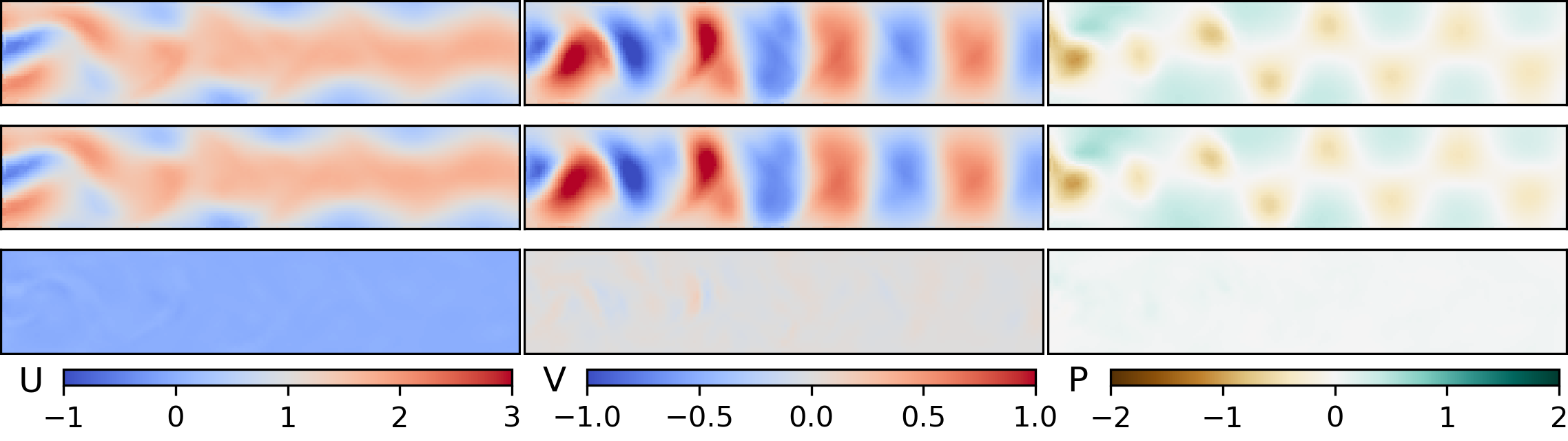}% train_15.0_5.5.csv/u_4650.png
  \caption{Full Component, data sampled at $t=4.65$ with $u_{inlet}=1.0$ and $d_y=0.11$}
  \label{fig:self}
  \end{subfigure}
  \begin{subfigure}[b]{0.8\textwidth}
  \centering
  \includegraphics[width=\textwidth]{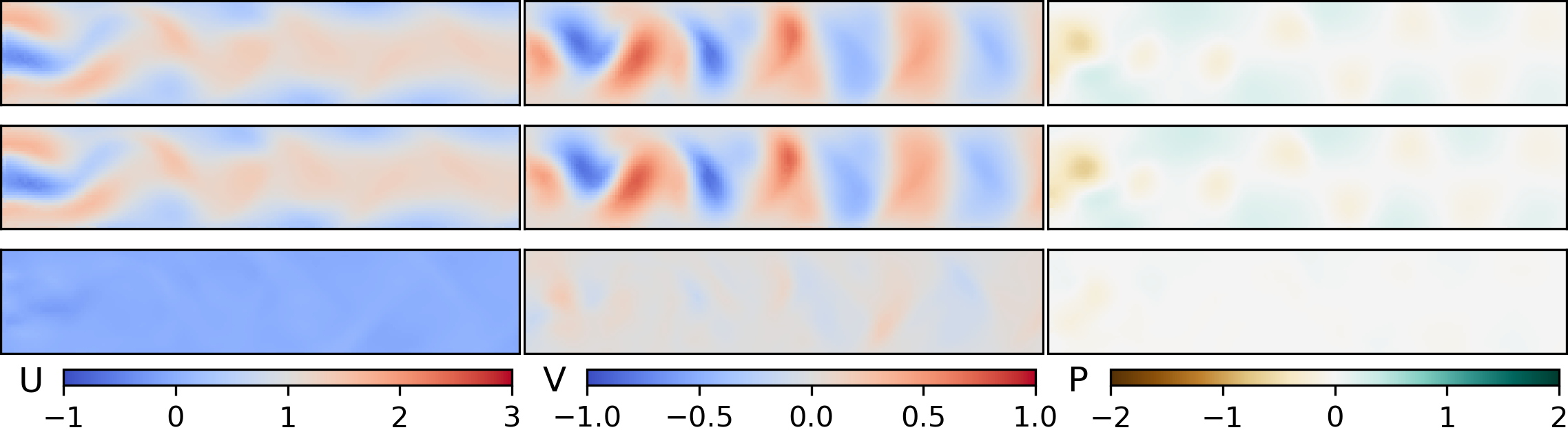}
  \caption{Full Component, data sampled at $t=3.95$ with $u_{inlet}=0.8$ and $d_y=0.11$}
  \label{fig:test}
  \end{subfigure}
  \caption{Visualization output as discussed in Section~\ref{sec:self} and \ref{sec:test}. In the same figure, the first row is the ground truth, the second row is the prediction and the third row is the error, other visualization figures have the same layout.}
  \label{fig:full-component}
\end{figure}

\subsubsection{Prediction within training geometry settings (Seen)}
\label{sec:self}
Figure~\ref{fig:self} demonstrates the ``worst'' prediction output (the one possessing the highest MSE among the instances) of our proposed framework when using the geometry in the training set ($t=4.65$, $u_{inlet}=1.0$ and $d=0.11$). It is clear that, within the training geometry settings, our proposed PINN framework provides accurate prediction on the inputs that do not appear at the training time stamps. The errors are barely noticeable on the visualization result and there is almost no phase shift.

\subsubsection{Prediction outside training geometry settings (Unseen)}
\label{sec:test}
Figure~\ref{fig:test} demonstrates the performance with the geometry settings that is unseen in the training set. We select the best (lowest MSE) prediction results and demonstrate it in Figure~\ref{fig:test}. We can conclude that our framework generally works well in terms of recovering the data distribution in the sample space -- or in terms of visualization, the shapes and artifacts in the images.

\subsubsection{Ablation Settings, Challenges and Discussion}
\label{sec:ablation}

\begin{figure}[ht]
  \centering
  \begin{subfigure}[b]{0.49\textwidth}
  \centering
  \includegraphics[width=\textwidth]{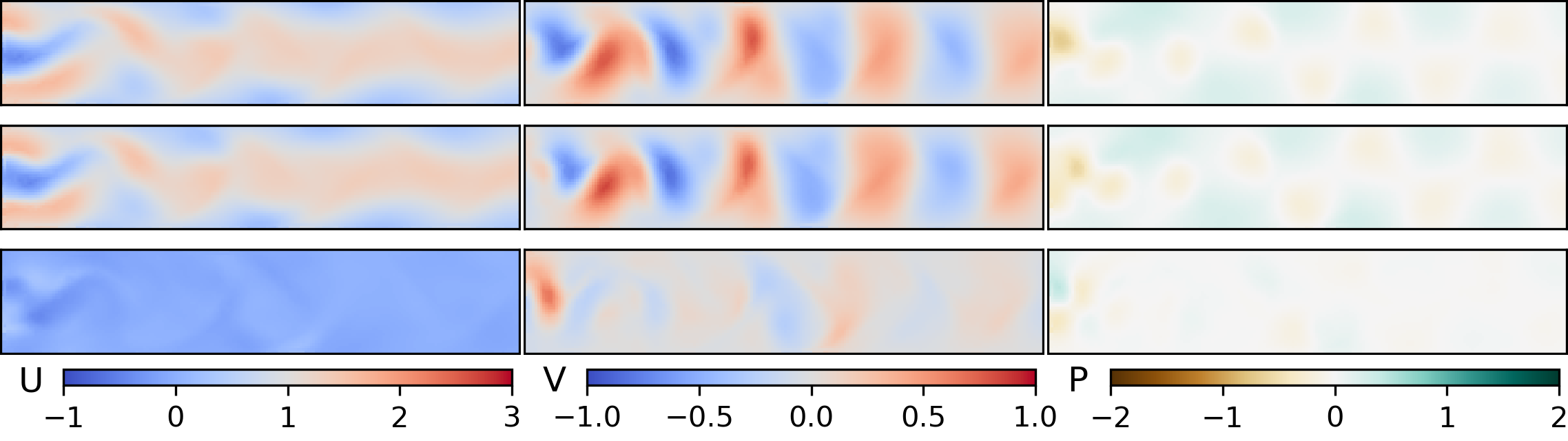}% train_15.0_5.5.csv/u_4650.png
  \caption{Prediction beyond time span, data sampled at $t=5.15$ with $u_{inlet}=0.8$ and $d_y=0.11$}%/test_12.0_5.5.csv/u_5150.png
  \label{fig:best_uncovered}
  \end{subfigure}
  \begin{subfigure}[b]{0.49\textwidth}
  \centering
 \includegraphics[width=\textwidth]{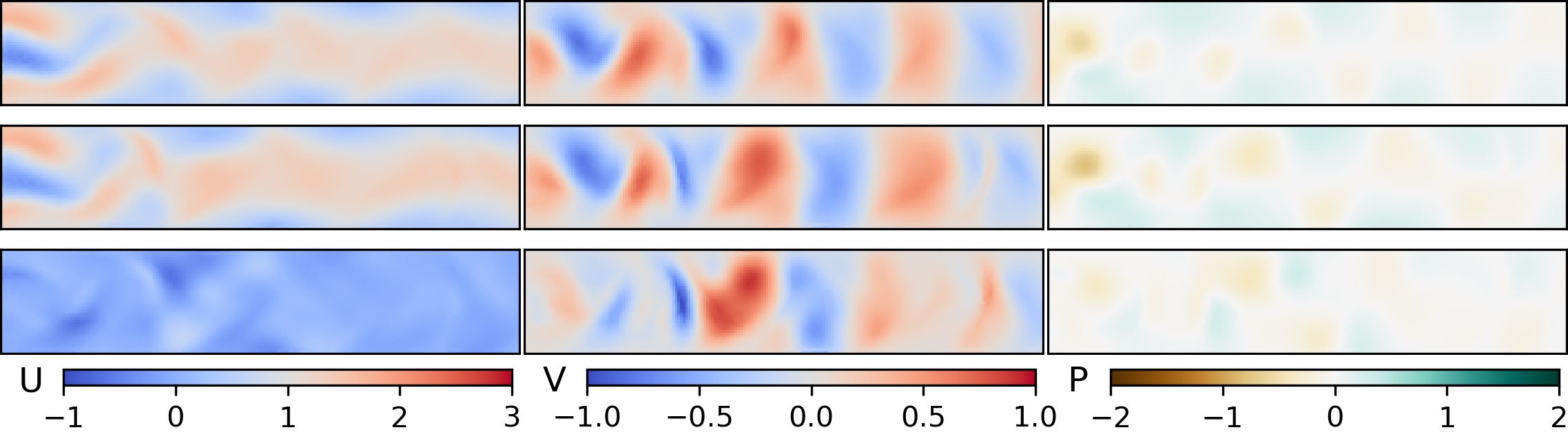}
  \caption{No-FFM, data sampled at $t=3.95$ with $u_{inlet}=0.8$ and $d_y=0.11$}%test_15.0_5.0.csv/u_3950.png
  \label{fig:no_fourier}
  \end{subfigure}
  \begin{subfigure}[b]{0.49\textwidth}
  \centering
  \includegraphics[width=\textwidth]{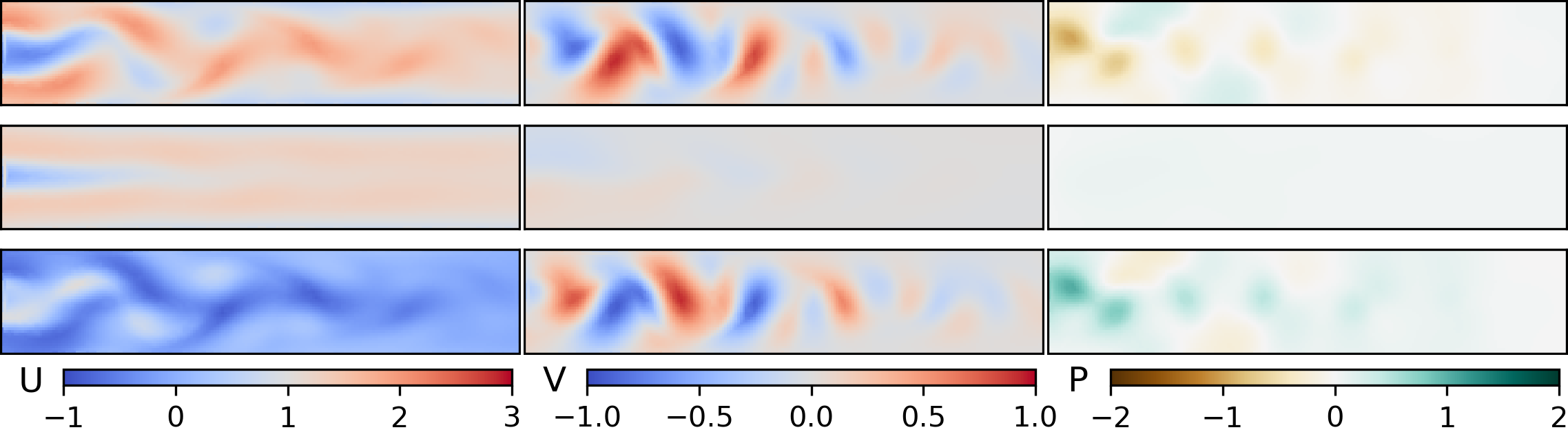}% test_15.0_5.0.csv/u_1750.png
  \caption{Strong-Reg, data sampled at $t=1.75$ with $u_{inlet}=1.0$ and $d_y=0.1$.}
  \label{fig:strong_reg}
  \end{subfigure}
  \begin{subfigure}[b]{0.49\textwidth}
  \centering
  \includegraphics[width=\textwidth]{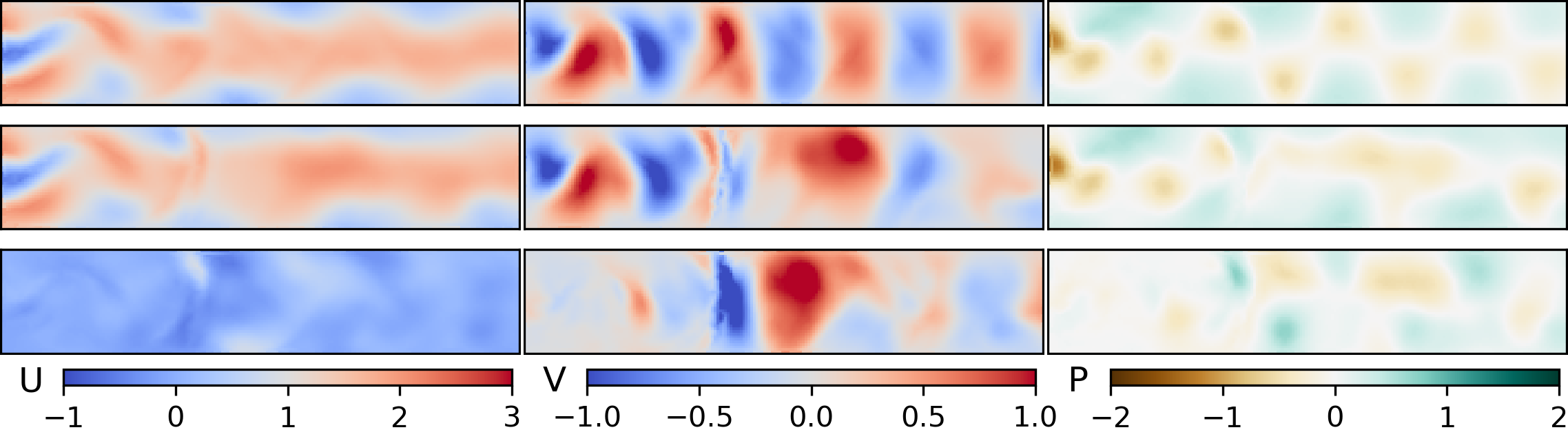}% 15.0_5.5 4750
  \caption{No-Reg/Overfit, data sampled at $t=4.75$ with $u_{inlet}=1$ and $d_y=0.11$}
  \label{fig:overfitting}
  \end{subfigure}
  \caption{Visualization results which are discussed in Section~\ref{sec:ablation}}.
  \label{fig:ablation}
\end{figure}

\paragraph{Beyond Time Span:} In Table~\ref{tab:loss_values}, we notice that points within time span $t\in[5,6]$ (denoted as ``Uncovered'') have lower performance or higher loss values. This is a major challenge in the proposed framework due to lack of further guidance in the training set and training process -- no sampled points are drawn from time span $t \in[5,6]$, However, as we illustrate the in Figure~\ref{fig:best_uncovered}, our proposed framework can still predict the flow velocity and pressure, albeit with some small phase shift.

\paragraph{Fourier Feature Mapping:} Figure~\ref{fig:no_fourier} demonstrate the visualization of a framework that removes the FFM component -- \textit{i.e.}, $x$, $y$, $t$ $u_{inlet}$ and $d$ were directly fed into the second MLP. We use the same geometry setting and time stamp as the one in Figure~\ref{fig:test}. From the output, we can conclude that, FFM provide crucial frequency and phase shift information in the output; and without it, the framework is not able to handle the complex frequencies and phase shifts among different combination of geometry settings.

\paragraph{Weighted Loss:} In most of the experiments, the loss weight parameter $\lambda$ was set as $0.001$, and Figure~\ref{fig:strong_reg} shows that a higher weighted PDE loss would inhibit the vorticity in the unseen settings. If we set $\lambda=0$, meaning that we do not introduce PDE regularization in the framework, we encounter over fitting as demonstrated in Figure~\ref{fig:overfitting}. 

\paragraph{Potential to Replace FeniCS-like Simulator:} When we conduct FeniCS simulation on one geometry setting, with a time step of $0.0001$ on a machine equipped with two Intel Xeon CPU E5-2620 v4 CPUs and 4 Nvidia Titan Xp, we spend more than $24,000$ seconds in simulating data over the time interval $[0s,6s]$. With our proposed framework, we only need $50$ time steps as anchor points from FeniCS, and within $15,000$ seconds (even we do not use early-stopping) we are able to acquire the same amount of data with comparable quality.
\section{Conclusion}
In this paper, we proposed a frequency-compensated PINN framework which can build high-accuracy surrogate for predicting simulation result in fluid dynamic design problems. In particular, we introduced and leveraged a Fourier feature mapping subnetwork to capture the periodicity present in the flow velocity and pressure; to conform with the underlying physics governed by the Strouhal number, this subnetwork learns the Fourier components as a functions of inlet velocity and cylinder size. Our results show that these Fourier features improve generalization in spatial and temporal domain as well for novel geometry settings. Future work would explore how this framework can be further extended to address fluid dynamic problems with more complex geometry and boundary conditions.

%%
%%
%%
%%%%%%%%%%%%%%%%%%%%%%%%%%%%%%%%%%%%%%%%%%%%%%%%%%%%%%%%%%%%%%%%%%%%%%%%%%%%%%%%%%%%%%%%%
%\clearpage
\small
\bibliography{PiML-4-Design--arXiv}
\bibliographystyle{abbrvnat}
%%%%%%%%%%%%%%%%%%%%%%%%%%%%%%%%%%%%%%%%%%%%%%%%%%%%%%%%%%%%%%%%%%%%%%%%%%%%%%%%%%%%%%%%%
%%
%%
%%
%\clearpage
\appendix
\section{Data Distribution}
Table~\ref{tab:data_distribution} demonstrates the distribution of the training, validation and test set. If a geometry setting appears in training set, we name it ``Seen'', otherwise ``Unseen''.

\begin{table}[ht]
  \centering
  \begin{tabular}{cccc}
    \toprule

    $\{u_{inlet}, d_y\}$ & Train & Validation & Test \\
    \midrule
    $\{0.8, 0.08\}$ & $57,700$ & / & $69,240$\\
    $\{0.8, 0.09\}$ & $57,500$ & / & $69,000$\\
    $\{0.8, 0.10\}$ & $57,400$ & / & $68,880$\\
    $\{0.8, 0.11\}$ & / & / & $137,760$\\
    $\{0.9, 0.08\}$ & $57,700$ & / & $69,240$\\
    $\{0.9, 0.09\}$ & / & $138,000$ & /\\
    $\{0.9, 0.10\}$ & $57,250$ & $68,700$ & /\\
    $\{0.9, 0.11\}$ & $57,400$ & / & $68,880$\\
    $\{1.0, 0.08\}$ & $57,500$ & / & $69,240$\\
    $\{1.0, 0.09\}$ & $57,400$ & / & $69,000$\\
    $\{1.0, 0.10\}$ & / & / & $137,400$\\
    $\{1.0, 0.11\}$ & $57,500$ & / & $68,880$\\
    \midrule
    Sum             & $517,350$ & $206,700$ & $827,520$\\
    \bottomrule
  \end{tabular}
  \caption{Statistics and split of training, validation and test sets. Due to internal mechanics of FeniCS, such as mesh grid settings, there are slight differences in the numbers of data points among different geometry settings.}
  \label{tab:data_distribution}
\end{table}

\end{document}